\documentclass[runningheads]{llncs}
\usepackage{graphicx}
\begin{document}
\pdfinfo{
	/Title (Segmentation-guided Domain Adaptation for Efficient Depth Completion)
	/Author (Fabian Märkert, Martin Sunkel, Anselm Haselhoff, Stefan Rudolph)
	/Subject (In this paper, we propose an efficient depth completion model based on a vgg05-like CNN architecture and propose a semi-supervised domain adaptation approach to transfer knowledge from synthetic to real world data to improve data-efficiency and reduce the need for a large database.)
	/Keywords (Deep Learning, Depth Completion, Domain Adaptation, Autonomous Drivings)
}

\title{Segmentation-guided Domain Adaptation for Efficient Depth Completion}
%
%\titlerunning{Abbreviated paper title}
% If the paper title is too long for the running head, you can set
% an abbreviated paper title here
%
\author{
	Fabian Märkert\inst{1} \and
	Martin Sunkel\inst{2} \and
	Anselm Haselhoff\inst{1} \and
	Stefan Rudolph\inst{2}
}
\authorrunning{F. Märkert et al.}
% First names are abbreviated in the running head.
% If there are more than two authors, 'et al.' is used.
%
\institute{
Ruhr West University of Applied Science, Bottrop, Germany\\
\email{fabian.maerkert@stud.hs-ruhrwest.de}\\
\email{anselm.haselhoff@hs-ruhrwest.de}
\and
e:fs TechHub GmbH, Gaimersheim, Germany
\email{\{martin.sunkel,stefan.rudolph\}@efs-techhub.com}\\}

\maketitle              % typeset the header of the contribution
\begin{abstract}
Complete depth information and efficient estimators have become vital ingredients in scene understanding for automated driving tasks. A major problem for LiDAR-based depth completion is the inefficient utilization of convolutions due to the lack of coherent information as provided by the sparse nature of uncorrelated LiDAR point clouds, which often leads to complex and resource-demanding networks. The problem is reinforced by the expensive aquisition of depth data for supervised training.

In this paper, we propose an efficient depth completion model based on a vgg05-like CNN architecture and propose a semi-supervised domain adaptation approach to transfer knowledge from synthetic to real world data to improve data-efficiency and reduce the need for a large database. In order to boost spatial coherence, we guide the learning process using segmentations as additional source of information.
The efficiency and accuracy of our approach is evaluated on the KITTI dataset. Our approach improves on previous efficient and low parameter state of the art approaches while having a noticeably lower computational footprint.

\end{abstract}
\section{Introduction}

% Warum Depth Completion
In automated driving typically multiple sensors, like camera, LiDAR and radar sensors, are used for a better understanding of a scene. Each of these sensors have their own advantages and drawbacks. A problem of LiDAR sensors is that the resolution is typically a fraction of the resolution of a camera image and in consequence depth information appears sparse and anisotropic. Convolutional neural networks are used to solve this problem by estimating dense depth maps in a task called depth completion.

% Problems and inefficiencies about Depth Completion and CNNs
However, a major problem of using convolutional neural networks for this type of task is the inefficiency of convolutions for sparse input data, which is typically solved by using complex networks with a high number of parameters. Due to the limited processing power available in autonomous vehicles, such networks are not suitable for real-time autonomy. Another problem with the task of depth completion is the expensive and hard to obtain ground truth data, which typically contains large unlabeled image regions (e.g. KITTI depth completion upper ~100px of the image). These unlabeled regions in the ground truth data lead to partly unrealistic depth estimates. In this paper we call these unlabeled regions out of training distribution regions.

% Kurzbeschreibung was wir machen wollen
In this paper, we build upon the VGG05-like ScaffNet \cite{learningtopology} for the task of depth completion. We first train the network on the synthetic Virtual KITTI \cite{vkitti} dataset and analyze its performance on real world data. We then apply a semi-supervised domain adaptation approach to transfer the synthetically learned topology to a real world domain. We improve the overall accuracy and out of training distribution sample capabilities by adding an additional segmentation input.

We analyze our approach in terms of accuracy for different configurations using standard metrics. The evaluation is performed for in and out of training distribution samples and finally we compare the results with a baseline network (ScaffNet) and the efficient state of the art FusionNet \cite{learningtopology}. In addition information about the computational footprint of the different approaches is provided. 

\section{Related Work}
The task of depth completion can be learned supervised using given ground truth training data or unsupervised using camera pairs or a series of images. 

In the past many approaches were based on matching stereo images or on the parallax effect and optical flow. These approaches were effective at the time but are now outdated due to the advancements in machine learning based methods.

Today's self-supervised high performing depth completion approaches are mostly based on photometric loss functions \cite{unsupervised_framework,learningtopology,visual_inertial_odometry}.
These approaches are useful and effective if no ground truth depth data is given but they lack performance in comparison to supervised approaches.

The authors of \cite{ddp} make use of synthetic datasets and unsupervised learning of depth priors but the approach is affected by a domain gap when applied to real world data. To overcome this problem \cite{domain_adaptation} applies advanced augmentation of synthetic dense depth maps and uses generative adversarial networks to close the domain gap between synthetic and real camera images. This way the domain gap is reduced but the training on the other hand became way more complex and resource-demanding.

The approaches proposed by the authors of \cite{PENNet} and \cite{non_local_spatial_propagation} deliver exceptional accuracies during supervised training, but are refining the estimated depth, which leads to complex, slow and resource-demanding networks.

Sparse depth maps are densified by applying triangular interpolation in the approach proposed by \cite{visual_inertial_odometry}. This allows for a more efficient use of convolutions but has the drawback of a high runtime due to the used CPU bound Delaunay triangulation algorithm.
Lately, the sparse input depth maps are densified using spatial pyramid pooling \cite{backprojection_layers,learningtopology}. This pooling method also allows for a more efficient use of convolutions and therefore allows for a smaller and more efficient network. The cited approaches can only marginally reduce the number of parameters and complexity due to the use of a secondary more resource demanding image encoder.

Depth estimation and semantic segmentation are a popular combination for multi-task learning networks \cite{UnsupervisedDepthSolvingDynamicObjectsMTL,singlestream} due to their similar network architectures, as well as their feature-sharing and performance improvement possibilities. These networks typically are vision based, perform a depth estimation and sometimes rely on resource-demanding refinement, which makes these type of networks inadequate for our use case.

\section{Efficient Depth Completion}
\begin{figure*}
	\centering
	\includegraphics[width=\textwidth]{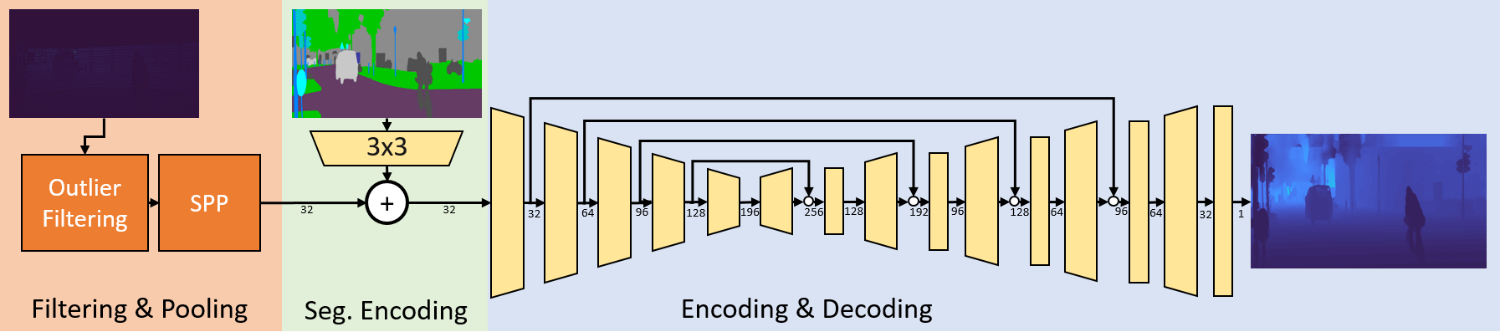}
	\caption{Network architecture, consisting of a filtering and pooling stage (Sec. \ref{sec:filter_and_pooling}), a segmentation encoding stage (Sec. \ref{sec:seg_encoding}) and an encoding and decoding stage (Sec. \ref{sec:encoder_decoder}) exemplary with a sample from KITTI dataset \cite{kitti}.}
	\label{fig:network}
\end{figure*}

In this section, we explain the overall architecture and components of our network, shown in Fig. \ref{fig:network}, which is based on the depth prior estimating ScaffNet proposed by the authors of \cite{learningtopology}. 

Our network consists of three stages. In the filter and pooling stage,  which is taken from the original ScaffNet, sparse depth maps are locally filtered to remove outliers and pooled using spatial pyramid pooling (SPP). The segmentation encoding stage, which we added, is used to capture additional object and shape information. For the final depth prediction a VGG05-like CNN is used in the encoding and decoding stage. In contrast to other depth completion approaches, we do not make use of additional camera images and thus are able to drastically decrease the number of parameters.

\subsection{Filtering and Pooling}
\label{sec:filter_and_pooling}
A major problem of CNNs for the task of depth completion is the inefficiency of convolutions for sparse input data. This inefficiency leads to complex and resource-demanding networks with a high number of parameters. Recent publications have shown that the inefficiency of convolutions on sparse input data can be solved by densifying the sparse depth maps before feeding them into the network \cite{visual_inertial_odometry,learningtopology,backprojection_layers}. 

For our network we use the same spatial pyramid pooling (SPP) as the ScaffNet. This spatial pyramid pooling approach uses max pooling kernels to densify the sparse depth maps without reducing their resolutions. This method has the advantage of being computationally inexpensive and allowing for an efficient use of convolutions, but has the drawback of being prone to sensor noise (e.g. reflections). To remedy this problem, we also use local outlier filtering before the spatial pyramid pooling. The filter removes depth values which exceed the minimum depth in the local neighbourhood of a pixel by a certain margin. For the implementation in this paper we chose 1.5 meters within a 7x7 neighbourhood.

\subsection{Segmentation Encoding}
\label{sec:seg_encoding}
Depth completion networks usually use camera images as an additional source of information. This improves the overall performance and theoretically allows for better out of training distribution sample capabilities, but comes at the cost of a higher computational footprint. % Due to image features that have to be extracted before being able to be used for depth completion

Due to not using any camera images in the ScaffNet, the object shapes are expected to be faulty and the out of training distribution sample capabilities are expected to be low. To improve the performance of the network, we add prior semantic scene information using given segmentation maps. We encode the segmentations by using a single 3x3 convolutional layer, which turns the one-hot encoded segmentation into a feature map with 32 channels, which is added to the output of the filtering and pooling stage.

\subsection{Encoder and Decoder}
\label{sec:encoder_decoder}
We adopt the VGG05-like encoder and decoder architectures from the ScaffNet. As shown in Fig. \ref{fig:network}, the encoder consists of 5 convolutional layers. Each layer has a stride of 2 and the padding is chosen to halve the resolution. For upsampling during decoding the nearest neighbour algorithm, followed by a 3x3 convolution is used. Concatenating skip connections from each encoding layer to the corresponding decoding layer are applied to allow for the reuse of low level features. The concatenated feature maps are fused using a 3x3 convolutional layer which halves the number of channels, but keeps the original resolution. We are using the ReLU activation function and apply batch normalization after each convolutional layer.

\section{Training and Domain Adaptation}
The training of the network consist of two phases. We start by training our network on the synthetic Virtual KITTI dataset \cite{vkitti} and then use the trained networks to adapt it to the real world KITTI dataset \cite{kitti} using different domain adaptation approaches.

\subsection{Synthetic Training}
The key advantages of synthetic datasets are the theoretically unlimited amount of training data and the availability of fully annotated dense depth maps. These dense depth maps allow for an overall more efficient training and for supervised training of out of training distribution regions. In addition the possible unlimited amount of data can allow for higher generalization capabilities.

To train the network using the synthetic dataset, the provided dense depth maps have to be reduced to sparse depth maps first. We use the technique proposed by \cite{learningtopology}, which applies the LiDAR scan pattern from the KITTI depth completion dataset to the synthetic dataset. The reduced sparse depth maps are used as an input for the network, while the provided dense depth maps are used as the ground truth.

We train the network using the normalized L1 loss

\begin{equation}
	l = \frac{1}{\left|\Omega \right|}\sum _{x\in \Omega }\left|\frac{d(x) - d_{gt}(x)}{d_{gt}(x)}\right|,
\end{equation}

where $\Omega$ denotes the pixel positions, $d$ the estimated depth and $d_{gt}$ the ground truth depth information.

\subsection{Training on Real Data}
A domain gap is expected due to the differences between the sparse depth maps from the synthetic dataset and the sparse depth maps from the real world dataset. Thus we need to adapt the network to the real world dataset. 

For the training on real world data we use a weighted L1 loss function
\begin{equation}
	l_{depth} = \frac{w_{depth}}{|\Omega|}\sum _{x\in \Omega }|d(x) - d_{gt}(x)|,
\end{equation} 
where $\Omega$ denotes the pixel positions with $d_{gt}(x) > 0$ and $w_{depth}$ dennotes a loss weight, which is defined later in Section \ref{implementation_details}.

\begin{figure*}
	\centering
	\includegraphics[width=1.0\linewidth]{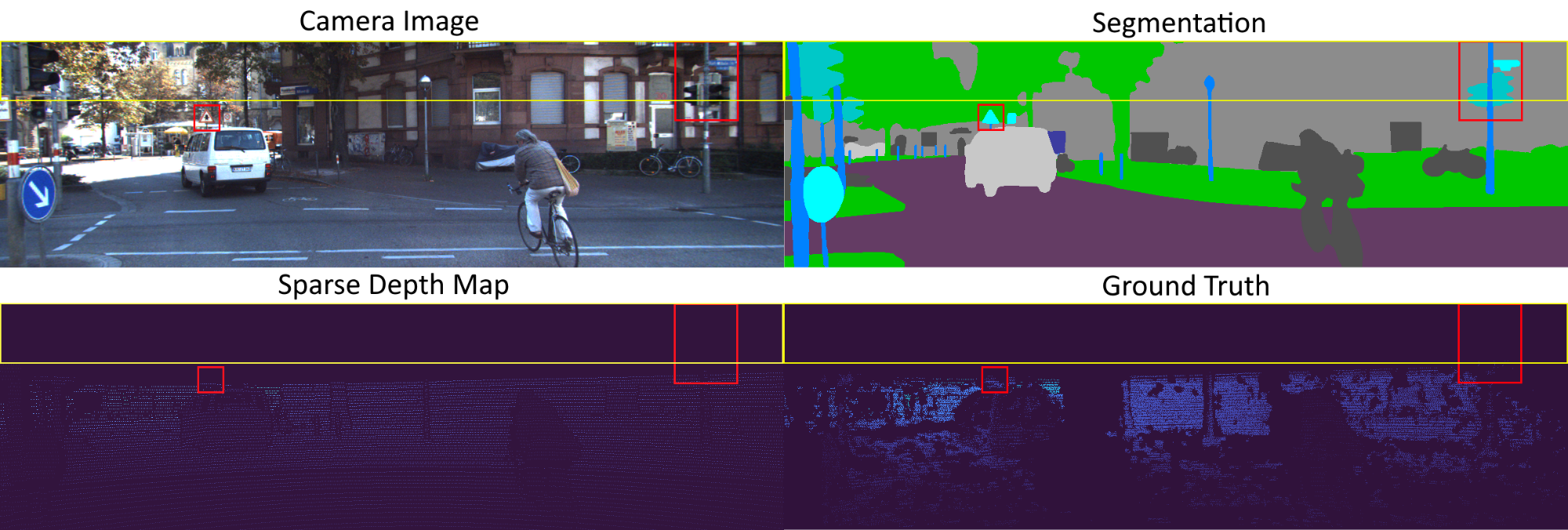}
	\caption{KITTI Validation Sample \cite{kitti}. The yellow box marks the out of training distribution region and the red boxes indicate regions of interests.}
	\label{fig:dataset_input_overview}
\end{figure*}

\subsection{Self-Supervised Domain Adaptation}
A problem with training the networks on real world depth data is that this data is typically sparse or semi-dense and may include regions completely without depth information. While sparse or semi-dense depth maps only leads to a decreased learning speed, regions completely without depth information lead to the loss of out of training distribution sample capabilities. The depth maps provided by the KITTI dataset do not contain any depth information for the upper third of the image. Thus, we expect the networks to lose their out of training distribution sample capabilities for these regions when training on real world data. To prevent this behavior, an adequate supervision for these regions is required.

The only option to provide supervision for these regions using the given data is to use a self-supervised photometric reprojection loss function. Such a loss function is commonly used in the field of self-supervised depth completion/estimation \cite{unsupervised_framework,backprojection_layers,learningtopology,visual_inertial_odometry}. It uses sequential camera images in combination with a known or estimated pose and the estimated depth to warp two sequential following camera images into each other and minimize the photometric discrepancy.

We adopt the loss functions used by \cite{learningtopology,unsupervised_framework} and the PoseNet proposed by \cite{learningtopology,visual_inertial_odometry}. The PoseNet estimates the 6D pose between the two sequential camera images and is learned indirectly by the minimization of the photometric discrepancy.

The resulting photometric reprojection loss function is given by
\begin{equation}
	l_{ph} = \frac{1}{2|\Omega|}\sum_{i\in(-1, 1)}\sum_{x\in \Omega}w_{co}|\hat{I}_{t+1}(x)-I_t(x)| +\\
	w_{ssim}(1-SSIM(\hat{I}_{t+1}(x), I_t(x))),
\end{equation}

with the pixel positions $\Omega$, where $\hat{I}_{t+i}>0$, the camera image $I_t$ at the timestamp $t$ and the inverse warped camera image $\hat{I}_{t+i}$ at timestamp $t+i$. The inverse warping is performed using the camera intrinsics $K$, the estimated pose $P_{t \rightarrow t+i}$ and a bilinear transformation 

\begin{equation}
	\hat{I}_{t+i} = bilinear(K P_{t\rightarrow t+i} d K^{-1} I_t).
\end{equation}

We also use the following edge aware smoothness loss
\begin{equation}
	l_{sm} = \frac{w_{sm}}{|\Omega|}\sum_{x\in \Omega }\lambda_x I_t(x)|\partial_x d(x)| + \lambda_y I_t(x)|\partial_y d(x)|
\end{equation} 
where $\lambda_xI_t(x) = e^{-|\partial_xI_t(x)|}$.

\section{Experimental Setup}
\label{experimantal}

First, we train the ScaffNet and our SegGuided approach on the Virtual KITTI dataset and evaluate them visually and using the metrics defined in \ref{metrics}.

We then training the networks on the real world KITTI dataset and apply three different domain adaptation approaches: 

1. (Supervised) We perform supervised training using the $l_{depth}$ loss and ground truth depth data. 

2. (Self-Supervised) We train the networks using the proposed loss functions, but only use the sparse input depth maps to compute the $l_{depth}$ loss.

3. (Semi-Supervised) We train the networks using the proposed loss functions and use the given ground truth depth maps to compute the $l_{depth}$ loss. 

We evaluate these approaches using performance and computational footprint metrics defined in \ref{metrics}. We also visually compare the estimated depth for out of training distribution regions to evaluate if our approach is able to maintain learned information from the training on the Virtual KITTI dataset.

\subsection{Dataset}
\label{dataset}
For the training of the networks on synthetic data, the Virtual KITTI dataset \cite{vkitti} is used.

For the training and evaluation of depth completion networks typically the KITTI depth completion dataset is used. This dataset contains 46375 pairs of sequential sparse and dense depth maps and camera images, but no segmentations, which are required for our approach. This makes this dataset unusable for our approach. The KITTI STEP dataset \cite{kitti_step} contains 8008 pairs of sequential camera images and segmentations, but no depth maps. For the training and evaluation of the networks we use the intersection of the KITTI depth completion and KITTI STEP dataset by matching the camera images. This leads to 7794 pairs of depth maps, segmentations and camera images. We used the KITTI Depth Completion training/validation data split and applied it to our intersection, which resulted in 5896 training and 1898 validation samples.

Due to different segmentation classes between the KITTI STEP and the Virtual KITTI dataset a mapping of the segmentation classes have to be performed beforehand.

\subsection{Metrics}
\label{metrics}
To evaluate the performance of the trained networks we use the Mean Absolute Error (MAE), Root Mean Squared Error (RMSE), inverse Mean Absolute Error (iMAE) and inverse Root Mean Squared Error (iRMSE).

As an indicator for the computational footprint we take a look at the number of parameters, the required inference time for a single sample and the required training time for a single training batch including loss computation and backpropagation. We do not make use of the required multiply add computations (MACs) because this metric can not cover the filtering and pooling stage and is hard to obtain. The amount of computations also have no indication about the inference performance because this metric does not take into account if these computations can be performed in parallel or need to be performed sequentially.

We visually compare the estimated depths for the out of training distribution regions, because the given ground truth data does not contain depth information for those regions. Therefore, the proposed metrics can not be used to evaluate these regions.

\subsection{Implementation Details}
\label{implementation_details}
We implemented the networks using PyTorch. A batch size of 8 has been used during training and the given depth maps and segmentations have been randomly horizontally flipped and cropped to a size of 768x320px. The Adam optimizer with a weight decay of 0.0001 has been used and a decaying learning rate of 0.00015 with a half-life of 50 epochs has been applied. 

During the self-supervised and semi-supervised training we weighted the loss functions with $w_{depth} = 0.05$, $w_{ssim} = 1.0$, $w_{co} = 0.2$ and $w_{smoothness} = 0.01$. 

We trained the networks for 300 epochs on an single GeForce RTX Titan V and used the checkpoint with the best validation loss for evaluation. During evaluation we performed a bottom crop to crop the depth maps and segmentations to a size of 1216x352px.

\section{Results}
We compare the performance of the networks by the presented metrics and the out of training distribution capabilities visually. We then compare the computational footprint by looking at the number of parameters as well as the required inference and batch training time.

\begin{table}
	\centering
	\caption{KITTI Validation performance results of the networks trained on the Virtual KITTI dataset}
	\begin{tabular}[h]{|c|c|c|c|c|}
		\hline
		Model 				& MAE			 		& RMSE					& iMAE				 	& iRMSE					\\
		\hline
		ScaffNet 			& \textbf{330.62} 		& \textbf{1505.47}		& \textbf{1.15}			& \textbf{3.92}			\\
		\hline
		SegGuided 		 	& 551.80 				& 2124.91 				& 6.96					& 706.59				\\
		\hline
	\end{tabular}
	\label{tab:results_vkitti}
\end{table}

\begin{figure}[t]
	\centering
	\includegraphics[width=0.85\linewidth]{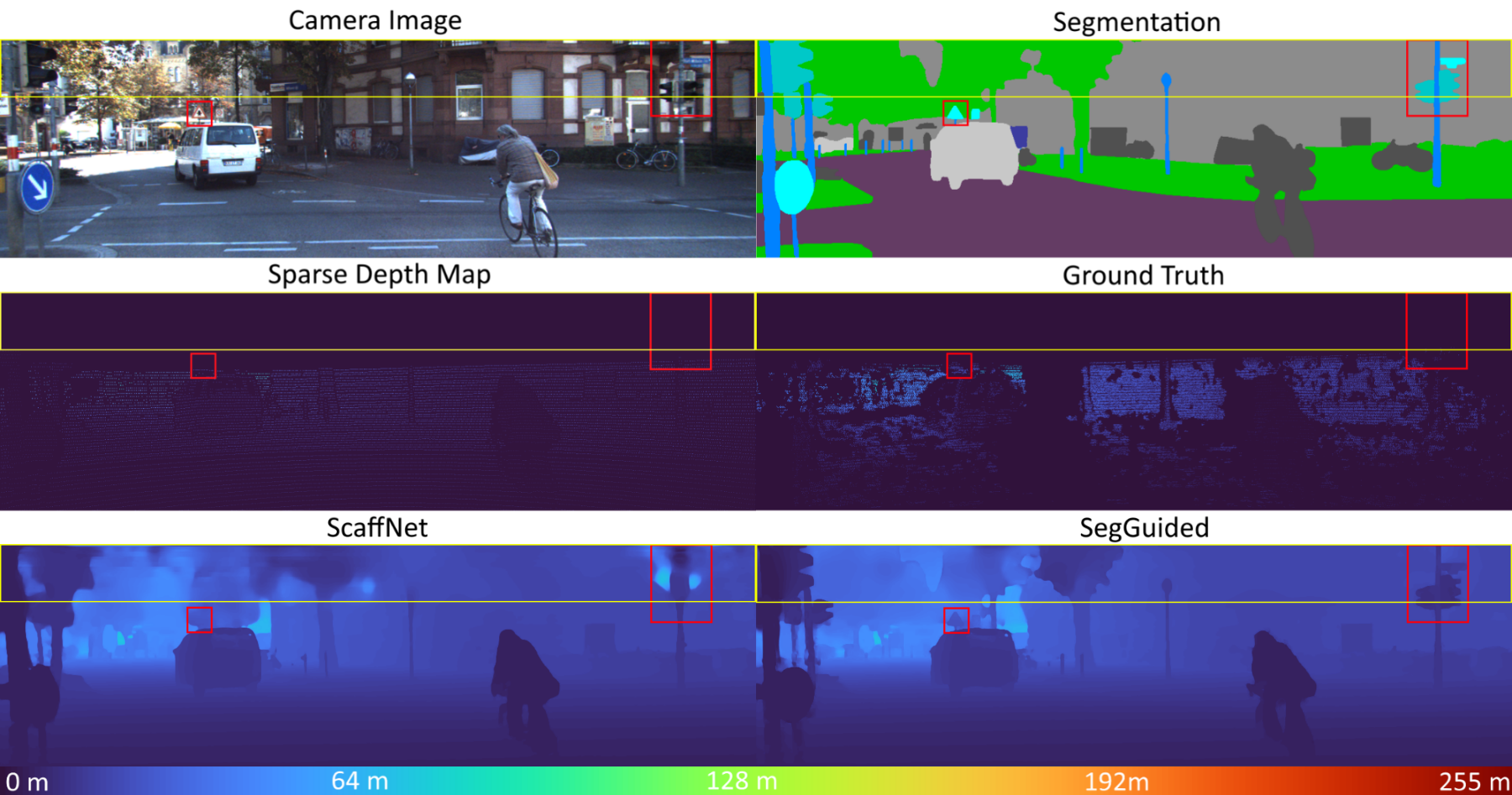}
	\caption{Visual KITTI Validation results of the networks trained on the Virtual KITTI dataset \cite{kitti}. The yellow box marks the out of training distribution region and the red boxes indicate regions of interests with big differences.}
	\label{fig:vkitti-comparison-colorbar}
\end{figure}

Table \ref{tab:results_vkitti} contains the results of our implementation for the ScaffNet and our SegGuided approach, which have been trained on the Virtual KITTI dataset and evaluated on the real world KITTI validation dataset. The performance of our SegGuided approach is lower than the performance of the ScaffNet, which is caused by unmatching segmentation object classes between the Virtual KITTI and KITTI STEP dataset. We tried to close this domain gap by remapping the object classes. We came to the result that the segmentations of these datasets are not perfectly matchable, which on the other hand allows for a better opportunity to show the effectiveness of our domain adaptation approach. As we expected Figure \ref{fig:vkitti-comparison-colorbar} shows improved out of training distribution capabilities and improved details for our SegGuided approach.

\begin{figure}[t]
	\centering
	\includegraphics[width=0.9\linewidth]{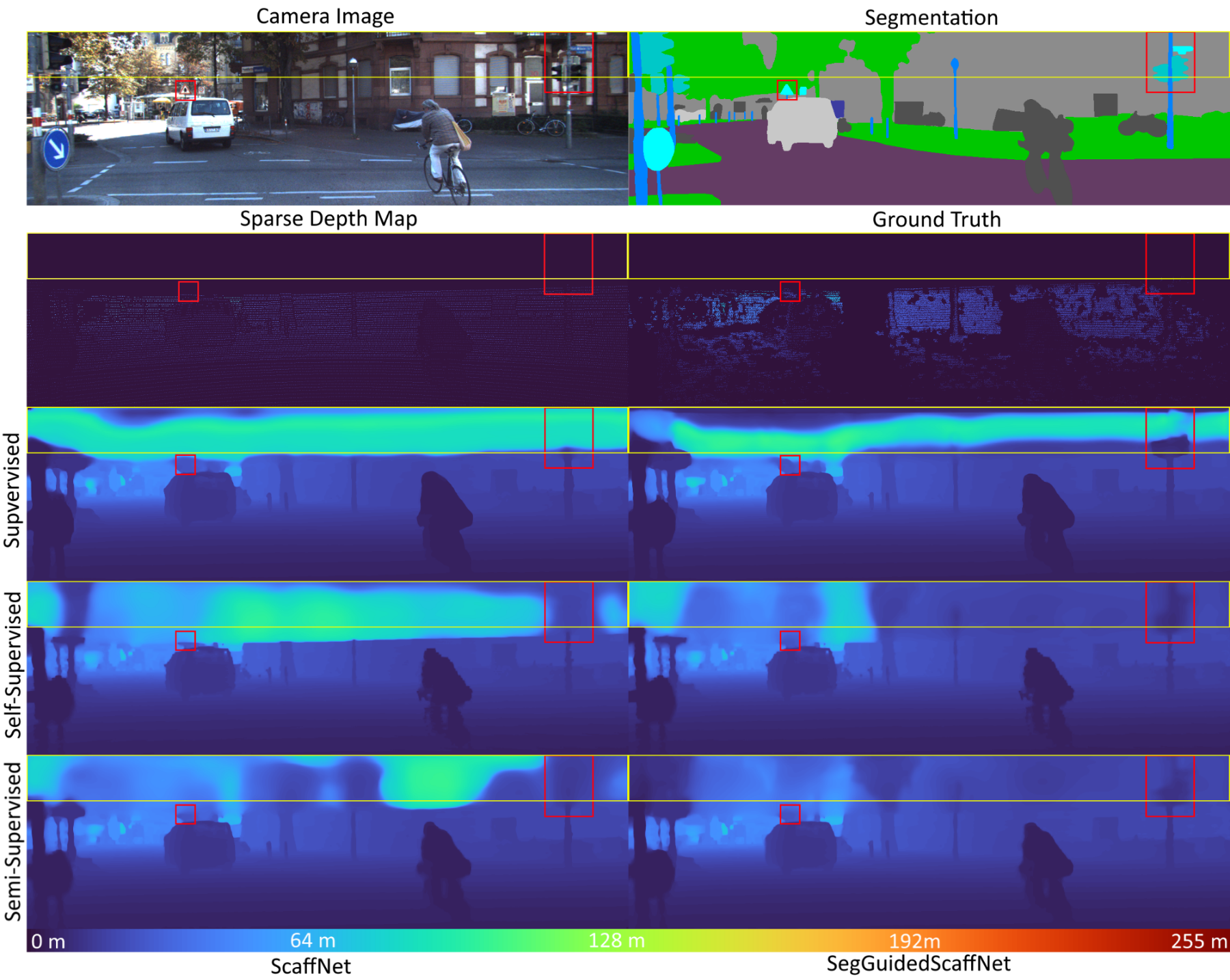}
	\caption{Visual KITTI Validation results of the networks trained on the KITTI dataset \cite{kitti} using our different domain adaptation approaches. Yellow boxes mark out of training distribution regions and the red boxes indicate regions of interests with big differences.}
	\label{fig:kitti-training-comparison}
\end{figure}

Table \ref{tab:results_retrained} shows the results of our different domain adaptation approaches. The supervised training of the ScaffNet shows minor improvements compared to the results in Table \ref{tab:results_vkitti}. Our SegGuided approach was able to adapt to the real world domain and shows an improved performance in comparison to the ScaffNet. When looking at Figure \ref{fig:kitti-training-comparison} the problem of the missing supervision of out of training distribution regions becomes visible. Both networks lost major parts of their generalization capabilities for the top region of the image and instead estimate a nearly constant depth. In a closer visual comparison our SegGuided approach has shown that it predicts a slightly more realistic depth for those regions.

\begin{table}
	\centering
	\caption{KITTI Validation results using our domain adaptation approache. The setup of our approach (Semi-Supervised) as well as the setup of the ablation studies (Supervised and Self-Supervised) are described in section \ref{experimantal}}
	\begin{tabular}[h]{|c|c|c|c|c|}
		\hline
		Model 						& MAE					& RMSE					& iMAE 					& iRMSE					\\
		\hline
		ScaffNet Supervised  		& 325.84 				& 1612.23 				& 1.08 					& 3.53 					\\
		\hline
		ScaffNet Self-Supervised  	& 411.53 				& 2579.26 				& 1.73 					& 4.63 					\\
		\hline
		ScaffNet Semi-Supervised 	& 283.34 				& 1269.58 				& 1.18 					& 3.71 					\\
		\hline
		SegGuided Supervised		& 309.70 				& 1477.40				& \textbf{1.05} 		& \textbf{3.12} 		\\
		\hline
		SegGuided Self-Supervised  	& 370.19 				& 1628.14				& 1.57 					& 4.65 					\\
		\hline
		SegGuided Semi-Supervised 	& \textbf{278.75} 		& \textbf{1146.78} 		& 1.30 					& 3.26 					\\
		\hline
		FusionNet 					& 289.57 				& 1202.91 				& 1.32 					& 3.62 					\\
		\hline
	\end{tabular}
	\label{tab:results_retrained}
\end{table}

The completely self-supervised training of both networks resulted in worse performances than the supervised trainings of them. Our SegGuided approach again shows a better performance than the ScaffNet. When comparing the results of the self-supervised ScaffNet visually to the results of the supervised ScaffNet, an improved but still unrealistic depth for out of training distribution regions can be observed. As seen on the traffic lights and the building on the right, our SegGuided approach predicts an improved and more realistic depth in those regions compared to the ScaffNet, but still predicts an unrealistic depth for the center top and the traffic lights on the left of the image.

Our Semi-Supervised domain adaptation approach shows performance improvements in comparison to the other two training approaches. 
Our SegGuided approach with this domain adaptation has the best MAE and RMSE of our compared networks while containing less than 20 percent of the parameters and requiring only about a third of the inference time of the previous efficient state of the art FusionNet, Table \ref{tab:computational_indicators}. The out of training distribution capabilities of this network were mostly able to be maintained. When comparing it to the results of our SegGuided approach from the training on the Virtual KITTI dataset, Figure \ref{fig:vkitti-comparison-colorbar}, the object shapes in the out of training distribution regions became blurred and small distant objects, like the traffic sign in the middle of the image, disappeared. The ScaffNet shows improved, but still partly unrealistic depth estimations for these regions compared to the results of the ScaffNet for the supervised and self-supervised training.

\begin{table}
	\centering
	\caption{Computational footprint indicators. The inference time is measured for a single sample and the training batch time is measured for a batch of 8, including loss calculation and backpropagation.}
	\begin{tabular}{|c|c|c|c|}
		\hline
		Model 			& Parameters  			& Inference			& Training Batch 	\\
		\hline
		ScaffNet		& \textbf{1,479,215} 	& \textbf{4.69 ms} 	& \textbf{172.15 ms}\\
		\hline
		SegGuided  		& 1,488,911				& 7.99 ms 			& 229.36 ms 		\\
		\hline
		FusionNet		& 7,825,569 			& 22.88 ms 			& 422.32 ms 		\\
		\hline
	\end{tabular}
	\label{tab:computational_indicators}
\end{table}

The reason why both of these networks improved their performances and out of training distribution capabilities in comparison to the other two domain adaptation approaches can be explained by the photometric loss function and the pose estimation. The used photometric loss function uses an additional PoseNet to estimate the pose and therefore learn the depth. Because of this correlation between pose and depth, the PoseNet can learn the pose when only a sparse depth loss is applied. We explain the improved performances and out of training distribution capabilities of our semi-supervised domain adaptation approach by a better learned pose due to the higher depth supervision of the ground truth depth maps compared to the used sparse depth maps used for the self-supervised training.

\section{Conclusion}
We adopted the VGG05-like ScaffNet depth completion network \cite{learningtopology} with a spatial pyramid pooling for sparse depth maps and added an input for segmentations to efficiently add object and shape information. 

We trained our approach and the original ScaffNet on the Virtual KITTI dataset and applied a semi-supervised domain adaptation approach to adapt the networks to the real world KITTI dataset. We compared the networks to each other and to the efficient state of the art FusionNet. We show the problem with out of training distribution regions by applying supervised and self-supervised training next to our semi-supervised domain adaptation approach.

We outperform the substantially larger FusionNet using the original ScaffNet and our domain adaptation approach while requiring only about 1.4M parameters and less than 5ms of inference time. With our SegGuided approach we noticeably improve the estimated depth for out of training distribution regions, which do not get covered by common evaluation metrics, while minimally increasing the number of parameters, increasing the inference time to about 8ms and requiring additional segmentations.

\section{Future Work}
For better estimations of out of training distribution regions additional scene informations are required. In our approach we use prior known segmentations. These segmentations are typically not directly given, but could be estimated by an additional segmentation network. Thus, we will train an additional segmentation network simultaneous with the depth completion network and use such pseudo segmentations to improve our depth completion network. To improve the performance and computational footprint of the segmentation and depth completion networks, we will be investigating feature-sharing mechanisms to create an efficient multi-task learning network.

\footnotetext{This document is licensed under the CC-BY-NC-SA-3.0 (https://creativecommons.org/licenses/by-nc-sa/3.0/legalcode).}
\end{document}